\documentclass[conference]{IEEEtran}
\IEEEoverridecommandlockouts
\usepackage{amsmath,amssymb,amsfonts}
\usepackage{algorithmic}
\usepackage{graphicx}
\usepackage{textcomp}
\usepackage{xcolor}
\usepackage{hyperref}

\usepackage[style=numeric,sorting=none,defernumbers=true]{biblatex}
\def\BibTeX{{\rm B\kern-.05em{\sc i\kern-.025em b}\kern-.08em
    T\kern-.1667em\lower.7ex\hbox{E}\kern-.125emX}}
\begin{document}
\makeatletter
\let\blx@rerun@biber\relax
\makeatother

\title{Impact of Network Topology on Byzantine Resilience in Decentralized Federated Learning
}

\author{
    \IEEEauthorblockN{Siddhartha Bhattacharya}
    \IEEEauthorblockA{\textit{Department of Computer Science} \\
    \textit{Michigan State University}\\
    East Lansing, USA\\
    bhatta70@msu.edu}
    \and
    \IEEEauthorblockN{Daniel Helo}
    \IEEEauthorblockA{\textit{Department of Computer Science} \\
    \textit{Michigan State University}\\
    East Lansing, MI \\
    helodani@msu.edu}
    \and
    \IEEEauthorblockN{Josh Siegel}
    \IEEEauthorblockA{\textit{Department of Computer Science} \\
    \textit{Michigan State University}\\
    East Lansing, MI \\
    jsiegel@msu.edu}

}

\maketitle

\begin{abstract}
Federated learning (FL) enables a collaborative environment for training machine learning models without sharing training data between users. This is typically achieved by aggregating model gradients on a central server. Decentralized federated learning is a rising paradigm that enables users to collaboratively train machine learning models in a peer-to-peer manner, without the need for a central aggregation server. However, before applying decentralized FL in real-world use training environments, nodes that deviate from the FL process (Byzantine nodes) must be considered when selecting an aggregation function. Recent research has focused on Byzantine-robust aggregation for client-server or fully connected networks, but has not yet evaluated such aggregation schemes for complex topologies possible with decentralized FL. Thus, the need for empirical evidence of Byzantine robustness in differing network topologies is evident. This work investigates the effects of state-of-the-art Byzantine-robust aggregation methods in complex, large-scale network structures. We find that state-of-the-art Byzantine robust aggregation strategies are not resilient within large non-fully connected networks. As such, our findings point the field towards the development of topology-aware aggregation schemes, especially necessary within the context of large scale real-world deployment. 
\end{abstract}


\section{Introduction}
As industry and public demand for increasingly compute-hungry and large models rise, concerns over the lack of computing accessibility and data centralization are also mounting \cite{foundation}. Edge computing has emerged as a strategy to address these pitfalls by bringing computation closer to data sources than typical cloud computing \cite{Verbraeken}. Federated learning (FL) brings the availability of machine learning models to edge devices. By participating in federated learning, nodes in an FL network can share their locally trained model to the FL network, and in return, they receive a global model that was trained on a diverse set of training data. Such a process appeals to the growing need for a machine learning paradigm which is privacy preserving, secure, and efficient for scenarios where training data is inherently distributed \cite{lili}. The deployment of federated learning in industry applications such as autonomous vehicles \cite{b5,b22}, agriculture \cite{fedcrop}, energy \cite{fedgrid}, cancer detection \cite{fedmed}, and other domains has created an additional demand for decision-making processes that leverage processes at the edge of the network \cite{b1}. 

The case of autonomous vehicles illustrates the potential of edge computing applied to machine learning for decision-making. Vehicles generate rich sources of training data that are inherently decentralized. Furthermore, the increasing automation of vehicles in recent years has increased the available on-vehicle computation. As such, federated learning has been used to collaboratively train machine learning models on vehicles for human activity recognition, image recognition, and localization \cite{b22}, enabling privacy preserving and scalable model training.

In centralized FL, the nodes share their locally trained model with a central parameter server, which aggregates node models and sends them back. Centralized FL is a straightforward training process, in which the node's local updates are sent to the parameter server, which aggregates these updates and sends them back. FedAvg \cite{mcmahan2023communicationefficient} is an effective communication-efficient algorithm to enable this aggregation step, especially in circumstances where participating nodes only have limited data for gradient computation \cite{b1}. Nonetheless, the utilization of a centralized server introduces its own set of complications. 

In decentralized FL, rather than sending model updates to a central aggregator, nodes share their models with neighbors in the network, in a peer-to-peer manner (P2P). The nodes individually perform aggregation of received models for each round \cite{dflsurvey}. This eliminates the risk of SPoF with a centralized parameter server, since each node has its own copy of the global model to be updated on each round. The P2P nature of decentralized FL also leads to lower communication overhead, since nodes are not required to be connected to a central server \cite{dflgos}. Therefore, the network topology can be configured in any manner.

However, there exists a major blind spot in the FL training process. Precisely, when the aggregation algorithm receives model updates from nodes, it cannot verify that these model updates were derived by honestly following the FL process. Nodes can construct arbitrary or malicious model updates that may mislead the global model, "poisoning" the training process. Such nodes are called Byzantine nodes \cite{byzstab}. The aggregator is then tasked not only with aggregating the received model updates but also ensuring that the process is robust to Byzantine nodes.

Several such aggregation functions have been proposed that claim to be robust to Byzantine attacks. These aggregation functions have been shown to protect the global model against various Byzantine attack strategies. However, a majority of prior work has primarily focused on Byzantine robustness in centralized FL \cite{b10, b11, liexp}. While some have explored Byzantine robustness in decentralized FL, these works have assumed a fully-connected topology \cite{bdfl, clippedgossip}, in which all nodes in the network share communication links with all other nodes, or an otherwise highly constrained topology \cite{clippedgossip}. However, such assumptions sideline one of the key motivations for decentralized FL, namely, the ability to have a dynamic network topology of participating nodes. Because decentralized FL operates as a P2P network, edge devices can be connected in various network topology configurations. In fact, this is the key motivator for edge devices such as automated vehicles, drones, and agricultural equipment to use decentralized FL, since they form distributed networks in which the network topology cannot be predetermined. To apply DFL to a large number of edge devices, networks with complex graphs must be considered. 

As such, \textbf{this paper investigates the following research question: How sensitive are complex decentralized FL networks to Byzantine attacks? We evaluate state-of-the-art Byzantine-robust aggregation strategies in a fully decentralized FL network under various network topologies with Byzantine nodes.}

\section{Relevant Background}
\subsection{Federated Learning}
Federated learning \cite{b7} is a distributed learning network in which a set of nodes collaboratively train a machine learning model on their local training data, typically by coordination with a central server \cite{b6}. In each round of federated learning, the nodes train locally for some number of epochs, then send their local model to the central aggregation server, which will perform some aggregation method to compute a global model. This global model is then sent back to the nodes. This process will repeat until convergence is reached. 

Centralized FL considers a set of $N$ nodes, each identically initialized with random model parameters $w$ and a local dataset $d_i$ sampled from some distribution $\mathcal{D}$. The network aims to find the optimal $w$ for the following optimization problem:
\begin{equation}
    \underset{w}{\text{min}} \; F(w) = \frac{1}{|N|}\sum_{i \in N}F_i(w, d_i)
\end{equation}
where $F_i$ is node $i$'s local loss function.

\subsection{Decentralized Federated Learning}
Centralized FL has attracted much attention from researchers, primarily in areas of training optimization and aggregation. However, centralized FL poses significant challenges to be applied in a practical edge network. Most of all, the over-reliance on the central aggregation server leads to a severe case of Single Point of Failure (SPoF) \cite{bbfl}. Furthermore, the mobility and spatial distribution of edge devices such as vehicles and personal electronics bring communication bandwidth and latency issues when communicating with a cloud server \cite{dflgos}.   

Decentralized FL offloads the tasks previously performed by the central server onto the training devices themselves in a P2P network. 

Let this network be represented by a graph $G(V,E)$ where $V := \{0,1, \ldots, n\}$ is the set of $N$ nodes and $E \in V \times V$ represents the directed edge set. Each edge $(i,j) \in E$ represents a connection between nodes $i$ and $j$. Let $V_i := \{j | (j,i) \in E$ denote the set of neighbors for each node $i$.

Instead of uploading their model updates to a central server for each round, nodes will instead share their model with neighboring nodes, $V_i$. Each node individually performs aggregation of the received models. The algorithm is given as follows
\begin{enumerate}
    \item {Nodes randomly initialized with $w_0$ for time $t=0$.}
    \item {Nodes train their local model, $w^i_t$ for $e$ epochs and compute model updates.}
    \item {Each node transmits $w^i_t$ to neighbors in $V_i$.}
    \item {Nodes perform some aggregation function, $\mathcal{A}(U_i)$ where $U_i := \{w^j : j \in V_i\}$ are the model updates from node $i$'s neighbors.}
    \item {Repeat steps 1-4 until convergence.}
\end{enumerate}

\subsection{Byzantine-Robust Aggregation}
Byzantine actors are defined to be nodes that deviate from the FL process and submit arbitrary or malicious model updates during FL training. The presence of a Byzantine actor poses a significant threat to the convergence of a global model in any FL network \cite{fedbyz}. While other attacks such as data manipulation attack the underlying distribution of the malicious node's training set, Byzantine agents are those who aim to disrupt the aggregation function of honest nodes. Precisely, a Byzantine agent, $j$ aims to disrupt $\mathcal{A}(U_i)$ for all $i$ such that $j \in V_i$. This is done by constructing $w^j_t$ so that the aggregation functions of node $j$'s honest neighbors output a global model that is incorrectly ``steered" by an arbitrarily constructed $w^j_t$.

Let $M \subset V$ be the set of Byzantine nodes and $M_i := \{m \in M | m \in V_i\}$ denote the set of Byzantine nodes in the neighborhood of node $i$. Then the objective of DFL in Byzantine networks now becomes

\begin{equation}
    \underset{w}{\text{min}} \; F(w) = \frac{1}{|V/M|}\sum_{i \in V/M}F_i(w, d_i)
\end{equation}

Thus, we seek an aggregation function which can can converge to a globally optimal model amongst the honest nodes, while being resilient to some number of Byzantine nodes. Federated Averaging (FedAvg) \cite{mcmahan2023communicationefficient} is a common aggregation scheme, where the global model is derived by taking the mean of all received model weight matrices. However, FedAvg is known to break down with just one Byzantine actor in the network \cite{b9}. As a result, a number of Byzantine-robust aggregation methods have been proposed to achieve Eq. (2) \cite{b10, b11, clippedgossip}. These methods assume an $F$\emph{-total} model \cite{b15}, which considers robustness to a total of $F$ Byzantine agents in the network, with a slight abuse of notation where $F=|M|$. 

Geometric-median (GeoMed) aggregation \cite{b10} finds $w$ and a set of weights $\{\alpha_j | j \in V_i\}$ which minimizes 
\begin{equation}
    \sum_{j\in V_i} \alpha_j||w - w^j_{t}||
\end{equation}
where $||\cdot||$ denotes the L2 norm in Euclidean space.

Krum aggregation \cite{b11} aims to select one of the received model updates with minimal Euclidean distance to a subset of $|V_i|-|M_i|-2$ updates:
\begin{equation}
    \text{Krum}\left(\{w^j | j \in V_i\}\right) = \underset{w}{\text{argmin}} \sum_{i \to j} ||w^i - w^j||^2
\end{equation}
where $i \to j$ denotes the set of $|V_i|-|M_i|-2$ neighboring model updates closest to node $i$ in terms of squared Euclidean distance. 

These methods, however, have assumed the setting of centralized FL, where all nodes are connected to a central server. While some works explore Byzantine-robustness in decentralized FL \cite{b10, b11, clippedgossip, b14}, they consider only highly constrained or fully connected networks. These works have not explored the inherent adaptability of DFL to dynamic network configurations or its ability to demonstrate scaling properties of networks.

\subsection{Network Models}
This paper explores Byzantine-robustness in two network models: Small-world network, and Scale-free network. We choose these network models because they have been commonly been found to naturally arise in real-world networks such as internet cable lines, social networks, and the structure of the US power grid \cite {b18,barabasi, Albert_2002}.

\begin{itemize}
\item The \textbf{small-world} network is generated according to the Watts-Strogatz model. This network characterizes the "small world" phenomenon with high clustering and short path lengths between nodes \cite{b18}. The topology resembles a ring lattice with some random connections going across the network. This network has been observed to emerge in real-world applications such as power grid communications.
\item The \textbf{scale-free} network is generated according to the Barabási–Albert model. This network demonstrates the property of preferential attachment \cite{Albert_2002}, or the increasing likelihood of new nodes in the network connecting to those with higher edge density. This model has been observed in social networks, Internet connections, airline networks, and others.
\end{itemize}

While \cite{b12} and \cite{b13} explored the effect of various network models on the convergence of DFL, Byzantine robustness in these networks has not been investigated. This work empirically evaluates the effectiveness of state-of-the-art Byzantine-robust aggregation schemes under these complex network topologies.

\section{Methodology}
\subsection{Decentralized FL Training}
This paper considers a DFL network of 128 nodes training a Convolutional Neural Network for handwritten digit recognition on the MNIST dataset. Similar image classification tasks and models have been applied to real-world computer vision applications, such as for traffic sign identification on autonomous vehicles \cite{trafficsign} and crop classification for agricultural devices \cite{cropclass}. 

In the implementation, each node has access to 250 distinct training samples. For each round of FL, the nodes train for 2 epochs, with a batch size of 32. Each node implements cross-entropy as the loss function and an Adam optimizer with a learning rate of $0.001$.

This experiment evaluates several aggregation schemes that have been proposed to be Byzantine-robust. The network is evaluated using Krum and GeoMed aggregation on each node. These functions have demonstrated resilience in centralized FL, but have yet to be tested in large-scale DFL networks. Our implementation follows that of \cite{fedstellar}.

\subsection{Network models}
The network is constructed with 128 nodes to simulate large training networks. This experiment considers small-world and scale-free networks under various configurations. 
\begin{itemize}
\item For the small-world network, let $k \in \{2,4,6,8\}$ determine the degree of each node to be 4 and $\beta = 0.2$ to be the probability of each connection to be rewired.
\item For the scale-free network, let $m_0 = 10$ be the number of hubs in the network which start with degree $m_0$. Subsequent nodes that are added to the network have degree $m_0$ and form connections according to preferential attachment.
\end{itemize}
\begin{figure}[htbp] 
    \centering
    \includegraphics[width=1\linewidth]{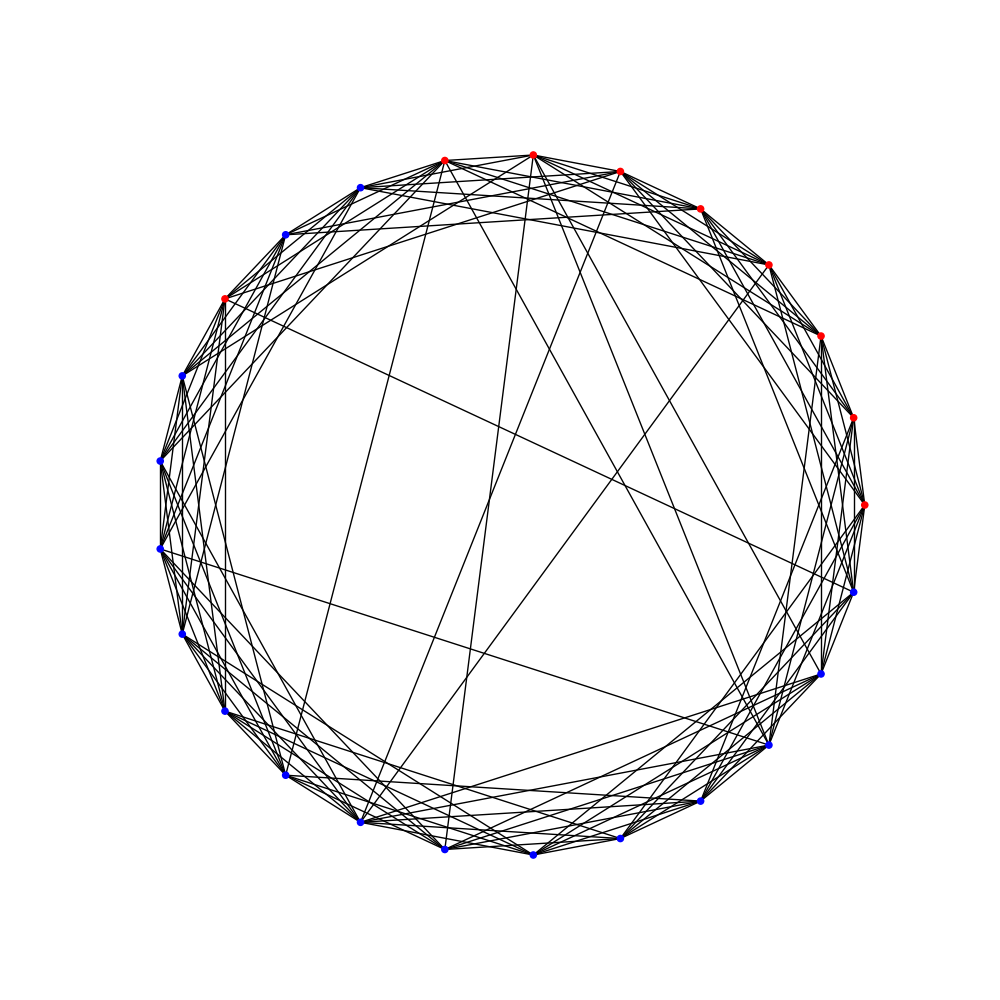} 
    \caption{Example of a small-world network, a ring-lattice structure with randomly rewired edges. The rewiring probability $\beta=0.1$.For each rewired edge, one end node is strategically positioned as Byzantine. The number of nodes is decreased from 128 for visualization.}
    \label{fig:centered}
\end{figure}

\begin{figure}[htbp] 
    \centering
    \includegraphics[width=1\linewidth]{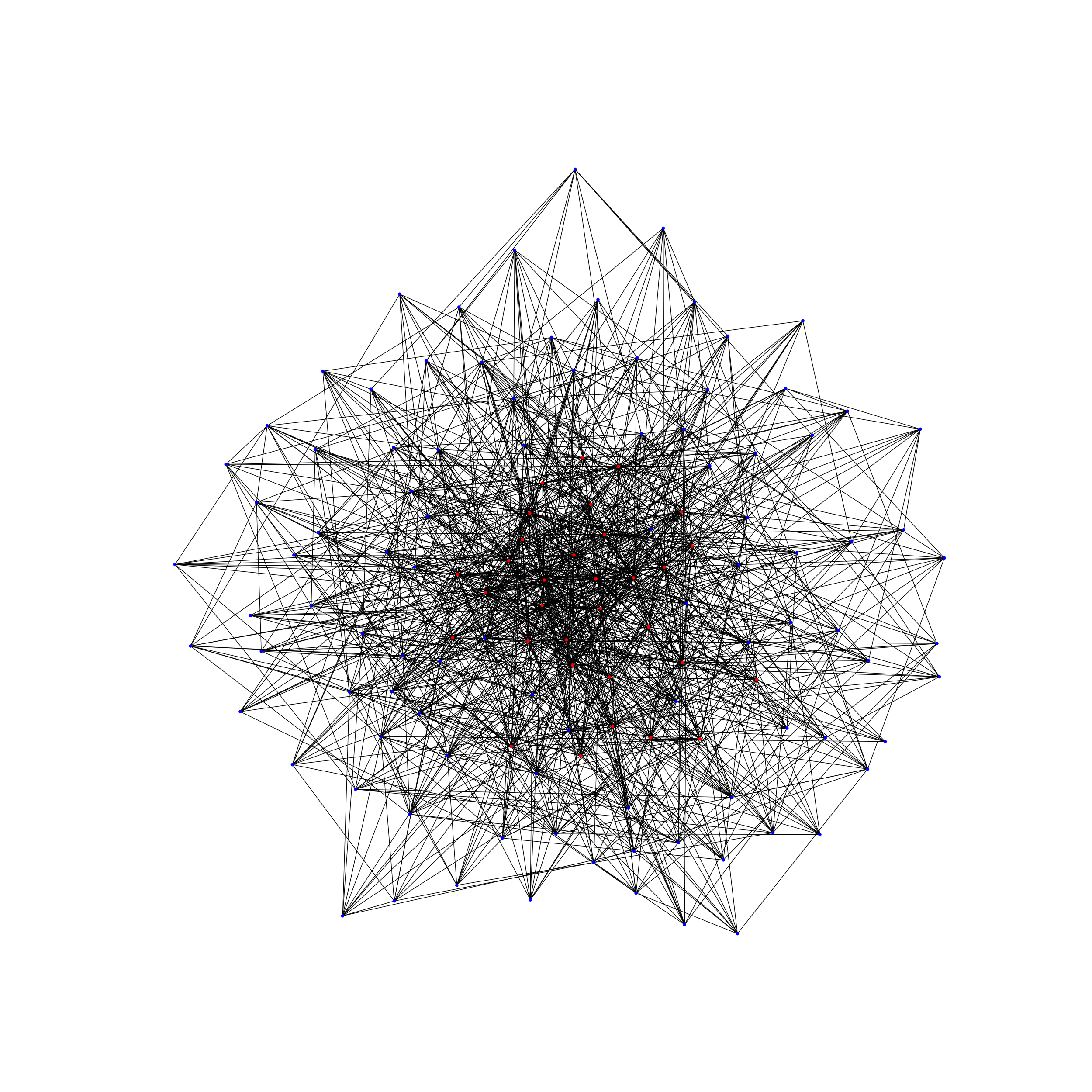} 
    \caption{Scale-free network with $n=128$ and   $m=10$ determines the degree of each node. The network grows according to preferential attachment, leading to hub nodes as seen in the center.}
    \label{fig:centered}
\end{figure}

\subsection{Byzantine Attacks}
This paper considers Byzantine Actors performing arbitrary model updates sampled from a Gaussian distribution. The malicious nodes randomly assign values to their model parameters from a normal distribution with a mean and standard deviation of 0 and 1, respectively.

This paper's motivation of evaluating arbitrary networks also motivates the evaluation of arbitrary Byzantine node placement. Random Byzantine node placement has been studied \cite{b19}, but under the assumption that such malicious nodes are dispersed randomly throughout the network. However, it is possible for a small group of collaborating Byzantine agents to strategically attack key positions within the network to take advantage of network structure \cite{b20}. Therefore, we evaluate both random and strategic placement of Byzantine nodes within the topology.

The selection of Byzantine nodes is controlled to evaluate the effect of placement and clustering of Byzantine actors. Let $p_i$ denote the probability that node $i$ is selected to be Byzantine and $P = \{p_0, p_1, \ldots p_n\}$. 

For the first configuration, the Byzantine nodes are placed randomly throughout the network, so that $p_0 = p_1 = \ldots p_n$. 
\textbf{Case 1: Random placement}. Byzantine actors are placed randomly throughout the network (each node has an equal probability of being Byzantine).\\
\textbf{Case 2: Strategic Placement}. Byzantine actors occupy strategic locations within the network. The details of these configurations for each proportion of nodes that are B are given as follows:
\begin{itemize}
\item \emph{Small-world network strategic Byzantine placement}. Every node with a rewired connection is chosen to be Byzantine. The number of rewired edges is varied by evaluating different values for $\beta \in \{0, 0.05, 0.1, 0.15\}$. Intuitively, this configuration is chosen because nodes with rewired edge connections have influence across larger parts of the network than those without rewired edges.
\item \emph{Scale-free network strategic Byzantine placement}. Let $S:=\{n_0, n_1, \ldots, n_k\}$ be the list of nodes sorted by degree in descending order. Then the top $b|S|$ nodes are selected to be Byzantine where $b \in \{0, 0.05, 0.15, 0.25\}$. In this case, Byzantine actors target the central hubs of the network that possess the most edges.
\end{itemize}

\section{Evaluation}
The system is evaluated at the end of each training round, using \textit{accuracy} as the metric to determine whether the network converged to an optimal global model. Accuracy is given by
\begin{equation}
    \frac{\text{\# of correct predictions}}{\text{\# of total predictions}}
\end{equation}
The accuracy of the system at the end of each round is the average accuracy of all \textit{honest} nodes.

Figure 3 demonstrates the system accuracy using Krum aggregation with randomly placed Byzantine nodes in scale-free and small-world networks. The baseline accuracy with no Byzantine nodes was 95\%. As the proportion of Byzantine nodes increased, the global accuracy decreased, dropping to 87\% under 30\% Byzantine nodes, and 12\% under 60\% Byzantine nodes. These results are nearly identical for both small-world and scale-free networks.

\begin{figure}[htbp] 
    \centering
    \includegraphics[width=1\linewidth]{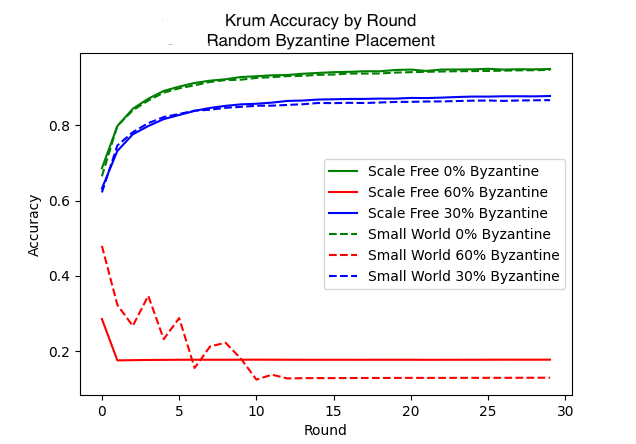} 
    \caption{Accuracy of Krum aggregation for randomly placed Byzantine nodes in both scale-free and small-world networks. Small-world $\beta=0.1$ and scale-free $m=10$.}
    \label{fig:centered}
\end{figure}

Figure 4 demonstrates system accuracy using GeoMed aggregation with randomly placed Byzantine nodes in scale-free and small-world networks. The final global model accuracy of these conditions is very similar to Figure 3, which used Krum, only that the scale-free network saw degradation (about 1\%) when considering 30\% Byzantine nodes.

\begin{figure}[htbp] 
    \centering
    \includegraphics[width=1\linewidth]{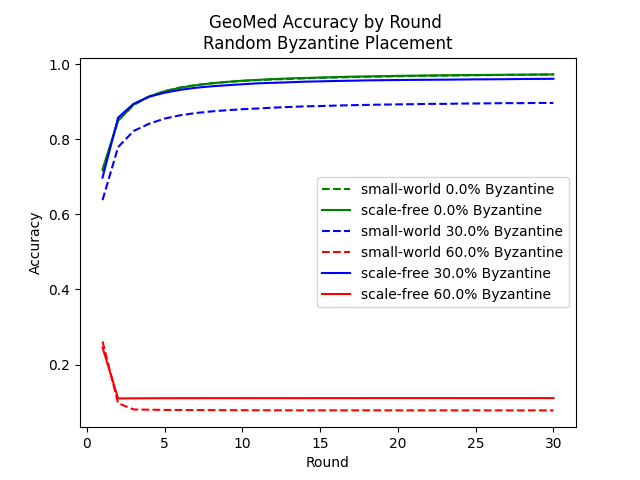} 
    \caption{Accuracy of Geomed aggregation for randomly placed Byzantine nodes in both scale-free and small-world networks. Small-world $\beta=0.1$ and scale-free $m=10$.}
    \label{fig:centered}
\end{figure}

Figure 5 describes accuracy over training rounds for strategic Byzantine placement within the small-world network for diverse values of $\beta$. The baseline used is $\beta=0.1$ with no Byzantine nodes. In this case, Krum and GeoMed converged to 94\% and 97\% accuracy, respectively. 

\begin{figure}[htbp] 
    \centering
    \includegraphics[width=1\linewidth]{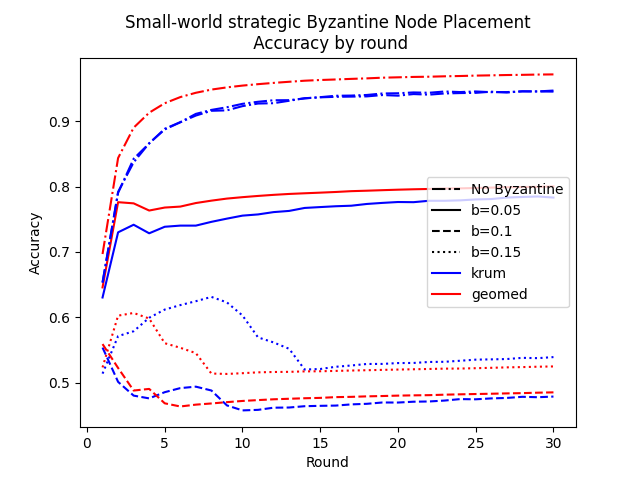} 
    \caption{Accuracy of Krum and GeoMed aggregation under strategically placed Byzantine nodes in small-world networks. For each $\beta \in \{0.05,0.1,0.15\}$, one end node of each rewired edge is set as Byzantine.}
    \label{fig:centered}
\end{figure}

Next, recall that for each rewired edge, one of the nodes corresponding to the edge is Byzantine. For $\beta = 0.05$, which corresponds to 27\% Byzantine nodes,  Krum and GeoMed both demonstrated degraded performance, with an accuracy of 78\% and 80\% respectively. Convergence breaks down for the network with $\beta=0.1$ and $\beta=0.15$, corresponding to Byzantine proportions of 50\% and 61\%, respectively. Figure 5 shows that for both values of $\beta$, the first 15 rounds of FL training show significant volatility in global model accuracy, until convergence at $\leq 50$\%.

Figure 6 describes accuracy over training rounds for strategic placement network configurations with diverse $\beta$ values under scale-free topology. Recall that in this case, $b$ refers to the proportion of top network hubs that are controlled by Byzantine nodes. In comparison to small-World, this topology is more robust overall to small numbers of Byzantine nodes. For $b=0.05$, Krum and GeoMed both retain the baseline accuracy of 95\% and 97\%, respectively. Both aggregation functions demonstrate degradation for $b=0.15$, both with an accuracy of 89\%. For $b=0.25$, both Krum and GeoMed diverge down to an accuracy of 5\% and 15\%, respectively.

\begin{figure}[htbp] 
    \centering
    \includegraphics[width=1\linewidth]{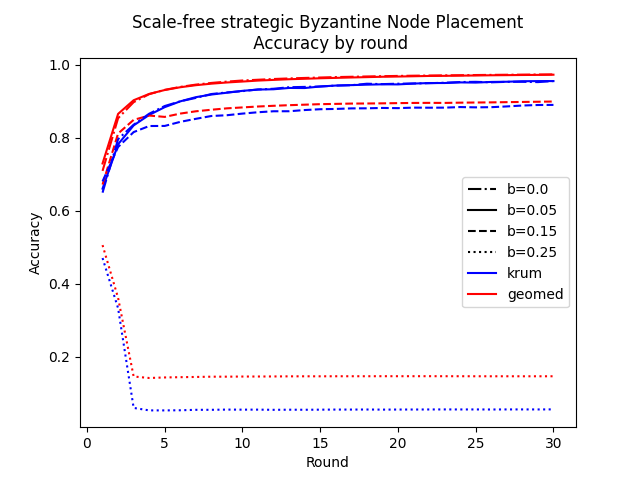} 
    \caption{Accuracy of Krum and GeoMed aggregation under strategically placed Byzantine nodes in scale-free networks. For each $b \in \{0.05,0.15,0.25\}$, the top $b$ proportion of nodes, sorted by degree, are set to be Byzantine.}
    \label{fig:centered}
\end{figure}

Our findings show, that small-world networks are more robust to strategic Byzantine nodes than scale-free. For the poisoned rewire connections with $\beta=0.05$, which corresponds to 27\% Byzantine nodes, the final model still exceeded 90\% accuracy. This can be explained by the uniform edge density of small-world networks. Even strategic Byzantine agents cannot overtake the overall connectivity of the network until they control a majority of the nodes. 

On the other hand, scale-free networks demonstrated poor resilience to strategic placement, failing to converge in similar proportions of Byzantine nodes. This is illustrated in Krum and GeoMed's robustness in scale-free networks to randomly placed 30\% Byzantine nodes, while both aggregation functions failed to converge a global model under only 25\% strategically placed Byzantine nodes. These findings show that the dense, centralized edge distribution of scale-free networks is more susceptible to Byzantine failure. Even a small number of poisoned hubs results in model divergence, since these nodes disproportionately control the network connectivity. 

As such, it is clear that strategically placed Byzantine actors in the network quantitatively lower robustness. The scale-free network's poor resilience to Byzantine nodes suggests that networks that form ``hubs" with higher degree than other nodes are vulnerable to strategic attack.

\section{Conclusion}
This study empirically evaluated the effect of network topology on the Byzantine resiliency of complex DFL networks under state-of-the-art aggregation schemes. While Byzantine-robust aggregation strategies have been developed for simple network models, they do not yet meet the challenge of complex DFL networks. For complex and large network structures, the two state of the art Byzantine-robust aggregation functions were not able to prevent model degradation for small numbers of strategically placed Byzantine agents. Such situations where small numbers of intelligent and network-aware Byzantine agents must be taken into account to prevent model divergence when implementing large-scale DFL networks. 

The limitation of this study is that it evaluated two Byzantine-robust aggregation functions and a simple noise attack, excluding several other sophisticated aggregation and attack strategies. However, the modern aggregation schemes follow the same assumption that the DFL network follows a client-server architecture or a fully connected topology. Furthermore, our results show that even a simple noise attack is effective in countering resilient aggregation functions in complex networks.

Our findings indicate that the assumption of fully connected networks must be relaxed for real-world DFL networks. This paper proposes a new direction of research for Byzantine-resilient DFL, namely, \textit{topology-aware} aggregation. These aggregation functions must be defined by the resiliency of the neighborhood of each node, rather than by resiliency to some quantity of Byzantine nodes irrespective of network topology. In future work, we aim to integrate theoretical bounds on resilient network topology from research in distributed optimization into the construction of DFL networks with robust aggregation functions.

\section{Acknowledgement}
We extend our thanks to Dr. Nitin Vaidya for his guidance in prior art in Byzantine-resilience, and to Uzair Mohammed and Avi Lochab in contributing to discussions. 
\section{appendix}
Code is available at \href{https://github.com/sidb70/DFL-Secure-Aggregation}{https://github.com/sidb70/DFL-Secure-Aggregation}
\printbibliography
\end{document}